\documentclass[journal,twocolumn,singlespace]{IEEEtran}
\usepackage{amsbsy,amsmath,amssymb,epsfig,bbm,mathrsfs,multirow,amsthm,subcaption,caption,graphicx,epstopdf,multicol,lipsum,float}
\usepackage[ruled,linesnumbered]{algorithm2e}
\usepackage{adjustbox}
\usepackage{multicol}
\usepackage[percent]{overpic}
\usepackage{enumerate}
\usepackage{tikz}
\usepackage[hidelinks]{hyperref}

\usetikzlibrary{arrows,calc}

\tikzstyle{line} = [draw, -latex']

\usepackage{lipsum}

\newenvironment{myalign}{\small\par\nobreak\noindent\align}{\endalign}
\begin{document}

\title{Re-route Package Pickup and Delivery Planning with Random Demands }
\author{\IEEEauthorblockN
{Suttinee Sawadsitang\IEEEauthorrefmark{1},
 Dusit Niyato\IEEEauthorrefmark{1},
 Kongrath Suankaewmanee\IEEEauthorrefmark{1},
 Puay Siew Tan\IEEEauthorrefmark{2}}
 \\
\IEEEauthorblockA{
\IEEEauthorrefmark{1} School of Computer Science and Engineering, Nanyang Technological University\\
\IEEEauthorrefmark{2}Singapore Institute of Manufacturing Technology (SIMTech) A*STAR }

}

\maketitle\thispagestyle{empty}

\begin{abstract} Recently, a higher competition in logistics business introduces new challenges to the vehicle routing problem (VRP). Re-route planning, also known as dynamic VRP, is one of the important challenges. The re-route planning has to be performed when new customers request for deliveries while the delivery vehicles, i.e., trucks, are serving other customers. While the re-route planning has been studied in the literature, most of the existing works do not consider different uncertainties. Therefore, in this paper, we propose two systems, i.e., (i) an offline package pickup and delivery planning with stochastic demands (PDPSD) and (ii) a re-route package pickup and delivery planning with stochastic demands (Re-route PDPSD). Accordingly, we formulate the PDPSD system as a two-stage stochastic optimization. We then extend the PDPSD system to the Re-route PDPSD system with a re-route algorithm. Furthermore, we evaluate performance of the proposed systems by using the dataset from Solomon Benchmark suite and a real data from a Singapore logistics 1company. The results show that the PDPSD system can achieve the lower cost than that of the baseline model. In addition, the Re-route PDPSD system can help the supplier efficiently and successfully to serve more customers while the trucks are already on the road. 
\end{abstract}


\begin{IEEEkeywords}
VRP, optimization, Stochastic, Re-route, Dynamic
\end{IEEEkeywords}

\section{Introduction}

In Singapore, the e-commerce industry is predicted to grow more than five times and reach S\$7.5 billion by 2026~\cite{ecommerce}. The e-commerce introduces complicated logistics requirements for package pickup and delivery. The vehicle routing problem (VRP) was first proposed in 1959~\cite{vrp} to help a shipper effectively plan its delivery. Many researchers have extended the traditional VRP in many aspects. One of the significant aspects is stochastic VRP, which can be referred to as VRP with one or more random parameters. The random parameters are common in a real situation in which shippers and suppliers do not have complete information about the delivery demand and other parameters a priori. Moreover, the planning by the shipper has to be dynamic as the customers can request for package pickup and delivery anytime. This is known as re-route planning issue in which the shipper must adjust and re-optimize truck utilization dynamically. Note that the planning becomes more challenging when the available trucks of the shipper cannot support all the requests, and the delivery has to be outsourced to a third-party carrier to minimize the customers' dissatisfaction.


In this paper, we consider the re-route planning of the shipper. In the scenario under consideration, the pickup and delivery requests from customers are generated dynamically, e.g., throughout the day. The shipper first decides whether to accept or reject the requests given available trucks. If the request is accepted, the shipper has to decide how to utilize and re-route the trucks that might be already serving other customers to accommodate the new request. On the other hand, if the request is rejected, the shipper will outsource the package pickup and/or delivery to a third-party carrier which typically incurs a higher cost. To achieve an efficient solution for the shipper, we propose a re-route package pickup and delivery planning with stochastic demand (Re-route PDPSD) system. The Re-route PDPSD system aims to help the shipper to (i) plan trucks' routing given that customers' package sizes are random, (ii) decide whether to accept or reject new customers while the trucks are serving the other customers, and (iii) re-plan the trucks' routing when the new customer requests are accepted. The main objective is to minimize the total delivery cost. In this regard, we first present the offline package pickup and delivery optimization with stochastic demand (PDPSD) system formulated as a two-stage stochastic optimization. The optimization can be solved as a linear programming. In the first stage, trucks are reserved in advance while the routes of trucks are decided in the second stage. Secondly, the Re-route PDPSD system is presented as an extension of the PDPSD system. In this case, a re-optimization algorithm is developed and included in the Re-route PDPSD system. To this end, we perform performance evaluation of the Re-route PDOSD system by using dataset from both Solomon benchmark suite and a real dataset obtained from Singapore logistics company. Compared with the offline version, the Re-route PDOSD system enables the shipper to serve more customers with a lower total delivery cost. 

\section{Related Work}

A number of researchers studied the Vehicle Routing Problem (VRP) to help a shipper effectively plan its delivery. The traditional VRP has been analyzed from different aspects, and the VRP with uncertainty is one of the major aspects~\cite{stochastic_survey}~\cite{dynamic_survey}. The VRP with uncertainty can be divided into two types, i.e., non-anticipative and anticipative. The former is only react to updates based on the new data, and the latter considers knowledge on the new data (e.g. probability) to predict the future~\cite{event-driven}. In general, the non-anticipative method is designed for a dynamic and deterministic problem, while the anticipative method is developed for a static and stochastic (also known as offline stochastic) problem. The difference between dynamic and stochastic are explained in Table~\ref{t_uncertainty}. Moreover, the survey~\cite{ds_survey} shows that few existing works consider both dynamic and stochastic problems simultaneously. Furthermore, almost all of them are dedicated for the VRP with one random parameter, e.g. customer location, customer demand, or traveling time. 

The authors in~\cite{ref_HH} proposed a model to solve the dynamic and stochastic VRP with a sample scenario hedging heuristic. They considered the case that customers' addresses and demands are known at the same time. They formulated the model as a multi-stage stochastic optimization. Each stage is separated from the other stages by the event that new requests are generated. In their system formulation, the first stage and the second stage decision variables are not decided at the same time. This means that the solution of the first stage is not based on the second stage. In their study, only the results from the heuristic solver were analyzed. Later on, the authors in~\cite{event-driven} considered the dynamic and stochastic VRP with random customers' locations and demands as in~\cite{ref_HH}.  The authors in~\cite{event-driven} proposed a new framework, which is java implementation of the multiple scenario approach. The adaptive variable neighborhood search is used in the framework. However, they only focused on the heuristic solver. In addition, the authors in~\cite{ref_container1} and~\cite{ref_container2} also focused on modeling and formulating the dynamic and stochastic optimization problem for logistics application. They obtained their experiment results from CPLEX solver. However, their main goal is on the container allocation, not vehicle routing. 

\begin{table}[t]
\caption{Classification of VRP with uncertainty}
\label{t_uncertainty}
\scriptsize
\begin{tabular}{c|cc}

& Deterministic input & Random inputs \\ \hline
Inputs are known in advance & static and deterministic & static and stochastic\\
Inputs change overtime & static and dynamic & dynamic and stochastic\\ 
\end{tabular}
\end{table}

Unlike the above works and other existing studies, we focus on defining reasonable scenarios and realistic system model as well as developing problem formulations, i.e., an exact method. We also consider both customers' locations and demands as the random parameters, where the values of both may or may not be known at the same time. As such, the system model is more suitable in practice, and the problem formulations are more general. Then, we propose the Re-route PDPSD system with two components, i.e., the stochastic optimization and the re-route algorithm. New dependency constraints and capacity constraints are proposed for pickup and delivery VRP. The optimization is formulated as a linear programming, which can be solved by standard solvers. 


\section{System Model and Assumptions}

\begin{figure}
\includegraphics[width=0.5\textwidth]{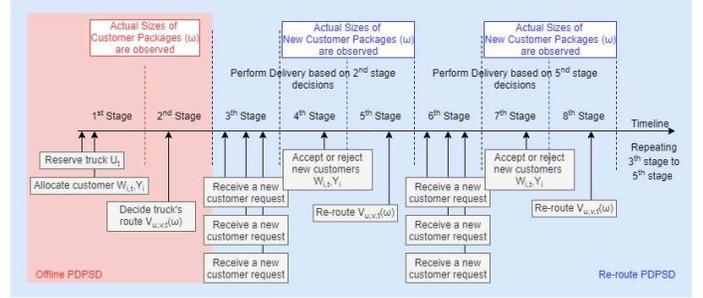}
\caption{Timing Diagram of the offline PDPSD system and the extended Re-route PDPSD system}
\label{fig_timeline}
\end{figure}

In this section, we present the system model for the re-route planning for a shipper. Then, we discuss the offline PDPSD and the Re-route PDPSD systems. The offline PDPSD system only considers the customers that submit requests before the first stage begins, i.e., a day before the delivery date. On the contrary, the Re-route PDPSD system allows the shipper to serve the customers that submit the requests later, i.e., when the trucks are already serving the other customers. Figure~\ref{fig_timeline} shows the timing diagram of both the systems. Again, the Re-route PDPSP system is an extension of the offline PDPSD system. In both the PDPSP systems, we consider two type of services, i.e., package pickup and delivery. The detail of each stage is explained as follows: 
\begin{itemize}
\small
\item \textbf{First stage:} Trucks are reserved in advance, and the customers are allocated to be served by either the shipper's truck or a third-party carrier. 
\item \textbf{Second stage:} After the sizes of customer packages are observed, the trucks' routes are decided. 
\item \textbf{Third stage:} Trucks perform delivery based on the routes from the second stage. Meanwhile, a new customer may request for a service, i.e., either pickup or delivery. 
\item \textbf{Fourth stage:} When the re-route/re-plan time is required, the system helps the shipper to decide whether to accept or reject customer requests. Note that the re-route/re-plan time needs to be set by the shipper. Meanwhile, the trucks still perform delivery on the road based on the routes from the second stage.
\item \textbf{Fifth stage:} After the actual sizes of customer packages are observed, the trucks' routes are re-generated by adding new customers to the plan and removing the served customers from the plan. 
\end{itemize}
Note also that the shipper can set multiple re-route time points by repeating the third, fourth, and fifth stages iteratively as shown in Figure~\ref{fig_timeline}.

In both the systems, we consider the problem that all trucks must start and end their journey at the shipper's depot. The depot is denoted as $\mathtt{D}$. Let $\mathcal{T}=\{\dots,T_t,\dots\}$ denote a set of trucks. Each truck has its capacity limit denoted as $l_t$. Let $\mathcal{C}=\{\dots,c_i,\dots\}$ denote a set of customers. Each customer may or may not require the service, i.e., pickup service and delivery service. If customer $c_i$ requires the service, then $k_i=1$, and $k_i=0$ otherwise. Without loss of generality, we consider that one customer has one package to be picked up or delivered. The package size of customer $i$ is denoted as $a_i$. Here, $a_i$ has a negative value if customer $i$ requires the delivery service, and $a_i$ has a positive value if customer $i$ requires the pickup service. In the first stage, the sizes of packages are unknown. Let $\Omega = \{\dots,\Omega_{\omega},\dots\}$ be a set of package size scenarios, where each $\omega$ consists of $a_i$ for all $i \in \mathcal{C}$, and thus $a_i(\omega)$ represents the size of package of customer $i$ when scenario $\omega$ occurs. 

When a customer requires the pickup or delivery service, the shipper can either serve the customer by a truck or outsource the customer to a carrier, i.e., pay penalty. Let $\mathcal{L} = \mathcal{C}\cup\{\mathtt{D}\}$ denote a set of locations. A dependency may exist between two locations in $\mathcal{L}$, where $u$ and $v$ are the indexes of set $\mathcal{L}$. For example, customer $i$ requires a truck to pick up a package, and this package needs to be delivered to customer $j$. As such, customer $i$ must be visited before customer $j$. Let $d_{u,v}$ denote a dependency input, where $d_{u,v}=1$ when location $u$ needs to be visited before and location $j$, and $d_{i,j}=0$ otherwise.


There are three costs involved in the systems. Let $\widehat{\mathfrak{C}}_t$ denote the initial cost of truck $t$, $\ddot{\mathfrak{C}}_{u,v}$ denote the routing cost from location $u$ to location $v$, and $\bar{\mathfrak{C}}$ denote the penalty or the outsourcing cost. 


\section{Optimization Formulations}
We present both the offline PDPSD and Re-route PDPSD formulations in this section. The offline and Re-route PDPSD systems are distinguished by the number of times that information is received. For the offline PDPSD system, the shipper plans the trip at the beginning of the day. By contrast, for the Re-route PDPSD system, the shipper can plan and re-plan package pickup and delivery of customers multiple times in a day. Let $\mathrm{T} = \{0,1,2,\dots\}$ denote a set of the request times that the shipper re-route/re-plans the trip in one day, and $\mathbb{T}$ denote the index of set $\mathrm{T}$. 
As shown in Figure~\ref{fig_timeline}, $\mathbb{T}=0$ represents the time before the first stage, $\mathbb{T}=1$  represents the third stage, $\mathbb{T}=2$ represents the sixth stage, and so on.
In the Re-route PDPSD system, the first/initial plan is decided based on the customers' information at time $\mathbb{T}=0$, and the system re-generates the new plan at time $\mathbb{T}=h$ based on (i) the previous trip at time $\mathbb{T}=r, r \leq h$ and (ii) the customer requests that the shipper receives before time $\mathbb{T}=h$. The offline PDPSD system considers only the request at time $\mathbb{T}=0$, while the Re-route system considers all the request times in set $\mathrm{T}$.




\subsection{Offline PDPSD system formulations}
\label{sec_off_stoc}

We formulate the offline PDPSD system as a stochastic programming. The objective is to minimize the total delivery cost, i.e., (i) the initial cost, (ii) the routing cost, and (iii) penalty or the outsourcing cost for handle the packages which are not served by a shipper's truck. The objective function is defined as in (\ref{eq_offline_deter_obj}).

In the PDPSD system, four binary, one integer, and one floating decision variables are defined. The details of the variables are given as follows: 
\begin{itemize}
\item $U_t$ is an indicator whether truck $t$ will be used or not. $U_t =1$ when truck $t$ will be used, and $U_t=0$ otherwise. 
\item $W_{i,t}$ is an indicator of the truck allocation. $W_{i,t}=1$ when customer $i$ will be served by truck $t$, and $W_{i,t}=0$ otherwise.
\item $V_{u,v,t}(\omega)$ is an indicator whether truck $t$ will use the path from location $u$ to location $v$ or not. $V_{u,v,t}=1$ when the path will be used, and $V_{u,v,t}=0$ otherwise.
\item $Y_i$ is an indicator whether customer $i$ will be served by the third-party carrier or not. $Y_i = 1$ when none of the shipper's trucks will serve customer $i$ and penalty or outsourcing cost needs to be paid, and $Y_i =0$ otherwise. 
\item $Q_i(\omega)$ is an auxiliary variable to ensure that the weight of all packages on the truck does not exceeds the capacity limit. 
\item $S_{i,t}(\omega)$ is an auxiliary variable for subtour elimination, i.e., $S_{i,t}(\omega) \in \{1,2,\dots,n\}$, where $n = \sum_{i\in\mathcal{C}}k_i$ is the total number of customer demands. 
\end{itemize}

\noindent Minimize: 
\begin{myalign}
\sum_{t \in \mathcal{T}}\widehat{\mathfrak{C}}_tU_{t}+ \sum_{i \in \mathcal{C}}\bar{\mathfrak{C}}Y_i + \sum_{t \in \mathcal{T}}\sum_{i,j \in \mathcal{L}}\sum_{\omega \in \Omega} P(\omega) \ddot{\mathfrak{C}}_{u,v}V_{u,v,t} (\omega)
\label{eq_offline_deter_obj}
\end{myalign}
Subject to: (\ref{eq_con_initial}) to (\ref{eq_con_capacity_limit}).
\begin{myalign}
&\sum_{i \in \mathcal{C}}W_{i,t} \leq \Delta U_{t},& \forall t \in \mathcal{T} \label{eq_con_initial}\\
& \sum_{t\in \mathcal{T}}W_{i,t} + Y_i =k_i, & \forall i\in \mathcal{C}\label{eq_con_allocation}\\
& \sum_{u\in \mathcal{L}}V_{i,u,t}(\omega) = k_iW_{i,t}, & \forall i \in \mathcal{C}, t \in \mathcal{T}, \omega \in \Omega \label{eq_con_route1}\\
& \sum_{u\in \mathcal{L}}V_{u,j,t}(\omega) = k_jW_{j,t}, & \forall j \in \mathcal{C}, t \in \mathcal{T}, \omega \in \Omega \label{eq_con_route2}\\
& \sum_{i \in \mathcal{C}} \Delta V_{\mathtt{D},i,t}(\omega)\geq \sum_{i \in \mathcal{C}}W_{i,t}, & \forall t \in \mathcal{T}, \omega \in \Omega \label{eq_con_route3}\\
& \sum_{i \in \mathcal{C}}V_{\mathtt{D},i,t}(\omega) \leq 1, & \forall t \in \mathcal{T}, \omega \in \Omega \label{eq_con_route4}
\end{myalign}
The constraint in~(\ref{eq_con_initial}) ensures that the initial cost must be paid when the truck is used. The constraint in~(\ref{eq_con_allocation}) ensures that a customers, who require the service, must be served by a shipper's truck. Otherwise, the shipper needs to pay an extra cost for the outsourcing. The constraints in~(\ref{eq_con_route1}) to~(\ref{eq_con_route4}) control the correctness of the routing path of each truck.

\begin{myalign}
& S_{i,t}(\omega) - S_{j,t}(\omega) + nV_{i,j,t}(\omega) \leq n-1, \forall i,j \in \mathcal{C}, t \in \mathcal{T}, \omega \in \Omega\label{eq_con_sub1}\\
&S_{i,t}(\omega) \geq W_{i,t}, \forall i \in \mathcal{C}, t \in \mathcal{T}, \omega \in \Omega \label{eq_con_sub2}\\
& d_{i,j}S_{i,t}(\omega) \leq S_{j,t}(\omega)-1, \forall i,j \in \mathcal{L}, t \in \mathcal{T}, \omega \in \Omega \label{eq_con_depen1}\\
&d_{i,j}Y_{i} \leq Y_{j}, \forall i,j \in \mathcal{C}\label{eq_con_depen2}\\
& d_{i,j} \leq Y_{i} - Y_{j} +1, \forall i,j \in \mathcal{C}\label{eq_con_depen3}\\
&d_{i,j}W_{i,t} \leq W_{j,t}, \forall i,j \in \mathcal{C}, t \in \mathcal{T}\label{eq_con_depen4}\\
& d_{i,j} \leq W_{i,t} - W_{j,t} +1, \forall i,j \in \mathcal{C}, t \in \mathcal{T}\label{eq_con_depen5}\\
&Q_{i,t}(\omega) + a_{j}(\omega) -Q_{j,t}\omega + \Delta V_{i,j,t}(\omega) \leq \Delta , \forall i,j \in \mathcal{C}, t \in \mathcal{T}, \omega \in \Omega \label{eq_con_capacity}\\
&q_{\mathtt{D}} + a_{i}(\omega) -Q_{i,t}\omega + \Delta V_{\mathtt{D},i,t}(\omega) \leq \Delta , \forall i \in \mathcal{C}, t \in \mathcal{T}, \omega \in \Omega\label{eq_con_capacity_st}\\
& Q_{i,t}(\omega) -l^{(c)}_{t} + \Delta W_{i,t} \leq \Delta, \forall i \in \mathcal{C}, t \in \mathcal{T}, \omega \in 
\Omega \label{eq_con_capacity_limit}
\end{myalign}
The constraints in~(\ref{eq_con_sub1}) and (\ref{eq_con_sub2}) eliminate a subtour from the solution. A subtour is a tour that consists of two or more unconnected routes in the solution. 
The constraint in~(\ref{eq_con_depen1}) ensures that the selected route must follow to the dependency input. For example, if $d_{i,j}=1$, then customer $i$ must be visited before customer $j$, $S_{i,t}(\omega) \leq S_{j,t}(\omega)$. The constraints in~(\ref{eq_con_depen2}) and~(\ref{eq_con_depen5}) ensure that when the dependency exists between customer $i$ and customer $j$, the shipper assignment for these two customers must be the same, i.e., either serving them by the same truck or outsourcing them to the third-party carrier. The constraints in (\ref{eq_con_capacity}) to (\ref{eq_con_capacity_limit}) ensure that the weight of all packages on a truck does not exceed the capacity limit of the truck. 

The above optimization can be solved as a linear programming. Let $\mathbf{U}_t$, $\mathbf{W}_{i,t}$, $\mathbf{V}_{u,v,t}(\omega)$, $\mathbf{Y}_i$, $\mathbf{Q}_i(\omega)$, and $\mathbf{S}_{i,t}(\omega)$ denote the optimized solutions of $U_t$, $W_{i,t}$, $V_{u,v,t}(\omega)$, $Y_i$, $Q_{i}(\omega)$, and $S_{i,t}(\omega)$ respectively. 

Next, the Re-route PDPSD system is discussed. We propose a new stochastic optimization formulation, which is dedicated to the Re-route PDPSD system together with the re-route algorithm.

\subsection{Re-route PDPSD system}
We then propose the Re-route PDPSD system with two components, i.e., (i) stochastic optimization formulation and (ii) re-route algorithm. The re-route algorithm calls the optimization as an inner function to re-plan the trip for the shipper. 

\subsubsection{Re-route Algorithm}

The re-route algorithm starts with solving the offline optimization problem in~\ref{sec_off_stoc}, as presented at Line 2 in Algorithm~\ref{al}. While the shipper's trucks travel to serve customers according to the plan, which is obtained from solving the offline PDPSD problem. New customers may request for the shipper service, which the shipper may know or may not know the package sizes. When the new customer requests for the service, the system automatically adds the customer and updates the customer detail as presented at line 5 to Line 8 of Algorithm~\ref{al}. Once it reaches the next re-plan time, the shipper observes (i) the actual customer package sizes, (ii) the current weights of the packages remaining in the trucks, (iii) the routes that the trucks visited and (iv) the current locations of the trucks (new origins), as presented at Line 9 to Line 16 of Algorithm~\ref{al}. Note that the shipper may set the re-planning time to be any instant, e.g. re-plan every two hours, re-plan at 12.00 am., or re-plan when the truck arrives at customer $i$. Let $\mathcal{O}^{\mathbb{T}}$ denote the origin of all trucks at time $\mathbb{T}$, i.e., $\mathcal{O}^{\mathbb{T}} = \{\mathtt{O}_{T_1},\mathtt{O}_{T_2},\dots\}$ and $\mathtt{O}_t$ denote the new origin of truck $t$, where $\mathtt{O}_t$ is the location of the last customer that truck $t$ visited. Let $\mathbf{q}_{\mathtt{O}_t}$ denote the weight of truck $t$ at origin $\mathtt{O}_t$. Let $\mathcal{F}$ denote a set of the depots and the served customers which does not include the last customer that trucks visited, i.e., $\mathtt{O}_t, \forall t \in \mathcal{T}$.  Let $\mathcal{C} ^{\mathbb{T}}$ denote a set of the customers that have not been served at time $\mathbb{T}$ and the origin location, i.e., $\mathcal{C}^{\mathbb{T}} =\mathcal{C}^{\mathbb{T}-1}\setminus \mathcal{F}$. Then, $ \{\mathtt{D}\}\cup\mathcal{C}^{\mathbb{T}}  = \mathcal{L}^{\mathbb{T}} $ , where $\mathcal{L}^{\mathbb{T}}$ is also the location that trucks can visit. Let $\mathcal{S}$ denote a set of only customers that need to be served, where $\mathcal{S} = \mathcal{C}^{\mathbb{T}} \setminus \mathcal{O}^{\mathbb{T}}$. After that, the Re-route PDPSD optimization problem, which is presented in the next section, is solved as linear programming by using the new parameters of time $\mathbb{T}$ as presented at Line 15 of Algorithm~\ref{al}. 

The algorithm will continue adding new customers and re-plan the trip by solving the online optimization problem until the shipper wants the system to terminate, e.g., at the end of the day. The solution is the combination of routes from the first plan to the last plan. An example of the solution can be found in Section~\ref{sec_solution}. The detail of the optimization formulation of the re-route PDPSD system is presented next.

\begin{algorithm}[]
\scriptsize
 
\SetAlgoLined
$\mathbb{T}=0$\\
Solve the optimization problem in (1). The values of decision variables (e.g. $\mathbf{U}_t$, $\mathbf{V}_{u,v,t}(\omega)$) are obtained as the solution at time $\mathbb{T}=0$.\\
\While{ Delivery time is not ended}{
 Pick up and deliver packages according to the plan at time $\mathbb{T}$\\
 \If{A new customer requests for the service}{
 Add the customer to sets $\mathcal{L}^{\mathbb{T}+1}$ and $\mathcal{C}^{\mathbb{T}+1}$\\
 Update customer detail, i.e., $k_{i}$, $a_{i}(\omega)$, $d_{i,j}$\\
 
 }
 \If{Shipper re-plans the trip}{
 Observe the actual visited customers ($\mathcal{F}$), and $\mathcal{C}^{\mathbb{T}+1} =\mathcal{C}^{\mathbb{T}}\setminus \mathcal{F}$\\
 Update the new origin ($\mathtt{O}_t \in \mathcal{O}^{\mathbb{T}+1}$ ) \\
 Update the current weight of the packages in the truck ($\mathbf{q}_t$)\\
 Observe the occurred scenario, i.e., $\omega'$, and use only $\mathbf{V}_{f,f',t} = \mathbf{V}_{f,f',t}(\omega')$ for the next plan \\
 $\mathbb{T} = \mathbb{T}+1$.\\
 Solve the optimization problem in (\ref{eq_online_sto}). The values of decision variables (e.g. $\mathbf{U}_t$, $\mathbf{V}_{u,v,t}(\omega)$) are obtained as the solution at time $\mathbb{T}$.\\}
}  
 \caption{Online optimization}
\label{al}
\end{algorithm}

\subsubsection{Stochastic Optimization Formulation for Re-route PDPSD system}
\label{sec_online_sto}


This optimization is an extension of the optimization in (\ref{eq_offline_deter_obj}). The objective function of the Re-route PDPSD optimization is presented in (\ref{eq_online_sto}), and it is subject to nine constraints similar to those of the offline PDPSD system. Again, we can obtain the value of $\mathbf{U}_t$, the value of $\mathbf{V}_{u,v,t}$, the value of $\mathbf{Y}_{f}$, set $\mathcal{O}$, set $\mathcal{S}$, and the value of $\mathbf{q}_{\mathtt{O}_t}$ from the previous plan. Nonetheless, instead of using $\mathcal{C}^{\mathbb{T}}$, $\mathcal{O}^{\mathbb{T}}$ and $\mathcal{L}^{\mathbb{T}}$, we use $\mathcal{C}$, $\mathcal{O}$ and $\mathcal{L}$ in the formulations as only one time $\mathbb{T}$ is considered at a time.


\noindent Minimize:
\begin{myalign}
&\sum_{t \in \mathcal{T}}
\widehat{\mathfrak{C}}_t\mathbf{U}_{t} +\sum_{t \in \mathcal{T}}\sum_{f,f' \in \mathcal{F}}\bar{\mathfrak{C}}_{f,f'}\mathbf{V}_{f,f',t} + \sum_{f \in \mathcal{F}}\ddot{\mathfrak{C}}\mathbf{Y}_f
\nonumber\\
&+\sum_{i \in \mathcal{C}}\ddot{\mathfrak{C}}Y_i
+ \sum_{\omega \in \Omega} \sum_{t \in \mathcal{T}}\sum_{\substack{u,v \in \mathcal{L}}} \mathbb{P}(\omega)\widehat{c}_{u,v}V_{u,v,t}(\omega),
\label{eq_online_sto}
\end{myalign}
subject to: (\ref{eq_con_allocation}), (\ref{eq_con_route1}), (\ref{eq_con_sub1}) to (\ref{eq_con_capacity}), (\ref{eq_con_capacity_limit}), and (\ref{eq_con_online_1}) to (\ref{eq_con_capacity_st_on}).


\begin{myalign}
&\sum_{i \in \mathcal{C}}W_{i,t} \leq \Delta \mathbf{U}_{t},& \forall t \in \mathcal{T} \label{eq_con_online_1}\\
& \sum_{j \in \mathcal{C}}V_{j,s,t}(\omega) = k_sW_{s,t}, & \forall s \in \mathcal{S}, t \in \mathcal{T},\omega \in \Omega \label{eq_con_online2}\\
& \sum_{i \in \mathcal{C}}V_{\mathtt{O}_t,i,t}(\omega) = 1, & \forall t \in \mathcal{T}, \omega \in \Omega\label{eq_con_online_3}\\
& \sum_{i \in \mathcal{C}}V_{i,\{\mathtt{D}\},t}(\omega) = 1, & \forall t \in \mathcal{T}, \omega \in \Omega \label{eq_con_online_4}
\end{myalign}
The constraints in (\ref{eq_con_online_1}) and (\ref{eq_con_capacity_st_on}) are adapted from the constraints in (\ref{eq_con_initial}) and (\ref{eq_con_capacity_st}), respectively, where the decision variable $U_t$ and $q_{\mathtt{D}}$ are replaced by new parameters $\mathbf{U}_{t}$ and $\mathbf{q}_{\mathtt{O}_t}$. The truck reservation $\mathbf{U}_{t}$ cannot be changed when the trucks already start the trip, and the trucks do not start their trips from the depot when time $\mathbb{T} \neq 0$. The trucks must start from the new origin ($\mathtt{O}_t$). The constraints in (\ref{eq_con_online2}) to (\ref{eq_con_online_4}) ensure the correctness of the route where truck $t$ departs the origin ($\mathtt{O}_t$), visits all customers, and returns to the depot ($\mathtt{D}$). 
\begin{myalign}
&\mathbf{q}_{\mathtt{O}_t} + a_{i}(\omega) -Q_{i,t}(\omega) + \Delta V_{\mathtt{\mathtt{O}_t},i,t}(\omega) \leq \Delta , \forall i \in \mathcal{C}, t \in \mathcal{T}, \omega \in \Omega\label{eq_con_capacity_st_on}
\end{myalign}
Similar to the offline PDPSD system, this optimization for Re-route PDPSD system can be solved as a linear programming problem. The optimization is called as a function inside Algorithm~\ref{al}.


\section{Performance Evaluation}

\subsection{Parameter Setting}
We consider the system with one truck. The truck belongs to shipper ($\ddot{c}_t=S\$0$ and $l_t = 50$). The routing cost of a truck is calculated as $\widehat{c}=0.1\times 1.05\times Distance $, where $0.1$ represents the average fuel consumption and $1.05$ represents the average fuel price in Singapore~\cite{ref_gov}. The package weights are varied for different experiments. The penalty of not serving a package or outsourcing a package to a carrier is set as $S\$16$ per packages, which is based to the cost of delivering 5 kilogram package by the Speedpost service offered by Singpost~\cite{ref_singpost}.   

We evaluate the system model with customer locations from the Singapore dataset and Solomon benchmark suite C101~\cite{ref_solomon}. We assume that all customers have demand, i.e., $k_i=1, \forall i \in \mathcal{C}$. The optimizations are implemented by using GAMS scripts and solved by the CPLEX solver~\cite{ref_gams}.

\subsection{A Solution Example}
\label{sec_solution}

\begin{figure}
\includegraphics[width=0.5\textwidth]{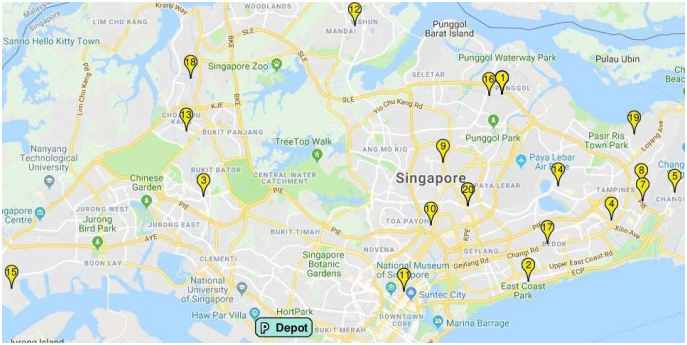}
\caption{Real Road Map}
\label{fig_map}
\end{figure}

We first present an example to help the readers understand the system clearly. We consider three plan and re-plan times per day (eight stages) in this example. At time $\mathbb{T}_0$, ten customers, i.e., customers $c_1$ to $c_{10}$, request for the service. At times $\mathbb{T}_1$ and $\mathbb{T}_2$, five customers request for the service, for which customers $c_{11}$ to $c_{15}$ submit the requests at $\mathbb{T}_1$, and customers $c_{16}$ to $c_{20}$ submit the requests at $\mathbb{T}_2$. The package weights of customers in $\mathbb{T}_0$ and $\mathbb{T}_2$ are known when the system receives the requests, where all packages at time $\mathbb{T}_0$ are set as $-5$ kilograms (package delivery requests), and all the packages at time $\mathbb{T}_2$ are set as $15$ kilograms (pickup requests). The requests at time $\mathbb{T}_1$ have two possible scenarios, which are $Q_{i}(\omega_1) = 5$ when $i = \{c_{11},c_{13},c_{15}\}$, $Q_{i}(\omega_1) = -5$ when $i = \{c_{12},c_{14}\}$, $Q_{i}(\omega_2) = 10$ when $i =\{c_{11},c_{13},c_{15}\}$, and $Q_{i}(\omega_1) = -10$ when $i = \{c_{12},c_{14}\}$. Customer $c_{11}$ must be visited before $c_{12}$, as well as customer $c_{13}$ and $c_{14}$, i.e., $D_{11,12}=1$ and $D_{13,14}=1$. Moreover, we set the occur scenario as $\omega_2$. The reason is that, in reality, we know the occur scenario before the fifth stage (refer to Figure~\ref{fig_timeline}). The locations of customers and the depot are presented in Figure~\ref{fig_map}. The solutions of each time $\mathbb{T}$ (loop iteration) are presented in Table~\ref{t_1}. The re-planning is done when the truck arrives at the third customer from the end of the route. According to Table~\ref{t_1}, the origin of the second plan is the location of customers $c_{10}$. Customers $c_3$ and $c_9$, which are planned to be served after customer $c_{10}$ in the first plan, are also included in the second plan. The actual route of the truck is shown in Table~\ref{t_1} with the total delivery cost of S\$80.715. 

\begin{table*}[]
\caption{The Solution Example of Section~\ref{sec_solution}}

\label{t_1}
\begin{tabular}{cccllcc}
\hline
\begin{tabular}[c]{@{}c@{}}Loop\\ Iteration\end{tabular}     & \begin{tabular}[c]{@{}c@{}}Starting\\ Weight\end{tabular} & Scenario           & \multicolumn{1}{c}{Routing Plan}                                                                 & \multicolumn{1}{c}{Outsourcing}         & \begin{tabular}[c]{@{}c@{}}Objective\\ Cost\end{tabular} & Distance       \\ \hline
1                                 & 50                            & $\omega_1$          & Depot$-c_2-c_4-c_5-c_7-c_8-c_1-c_6-c_{10}-c_3-c_9-$Depot                                                     &                         & 10.028                          & 95.5         \\ \hline
\multirow{2}{*}{2}                        & \multirow{2}{*}{10}                    & $\omega_1$          & $c_{10}-c_{15}-c_{11}-c_{3}-c_{9}-c_{13}-c_{12}-c_{14}-$Depot                                                   &                         & \multirow{2}{*}{10.290}                 & \multirow{2}{*}{98.0} \\
                                 &                              & $\omega_2$          & $c_{10}-c_{15}-c_{11}-c_{3}-c_{9}-c_{13}-c_{12}-c_{14}-$Depot                                                   &                         &                             &            \\ \hline
3                                 & 30                            & $\omega_1$          & $c_{13}-c_{18}-c_{12}-c_{16}-c_{14}-$Depot                                                            & $c_{15}$, $c_{17}$, $c_{19}$, $c_{20}$     & 70.678                          & 63.6         \\\hline
\textbf{\begin{tabular}[c]{@{}c@{}}Actual\\ Route\end{tabular}} & \textbf{50}                        & \multicolumn{1}{l}{\textbf{}} & \textbf{\begin{tabular}[c]{@{}l@{}}Depot$-c_2-c_4-c_5-c_7-c_8-c_1-c_6-c_{10}-c_{15}-c_{11}-$\\ $c_{3}-c_{9}-c_{13}-c_{18}-c_{12}-c_{16}-c_{14}-$Depot\end{tabular}} & \textbf{$c_{15}$, $c_{17}$, $c_{19}$, $c_{20}$} & \textbf{80.715}                     & \textbf{159.2}    \\\hline
\end{tabular}
\end{table*}

\begin{figure*}[t]
\hspace{-2em}\begin{minipage}{0.4\textwidth}
\begin{overpic}[width=\textwidth]{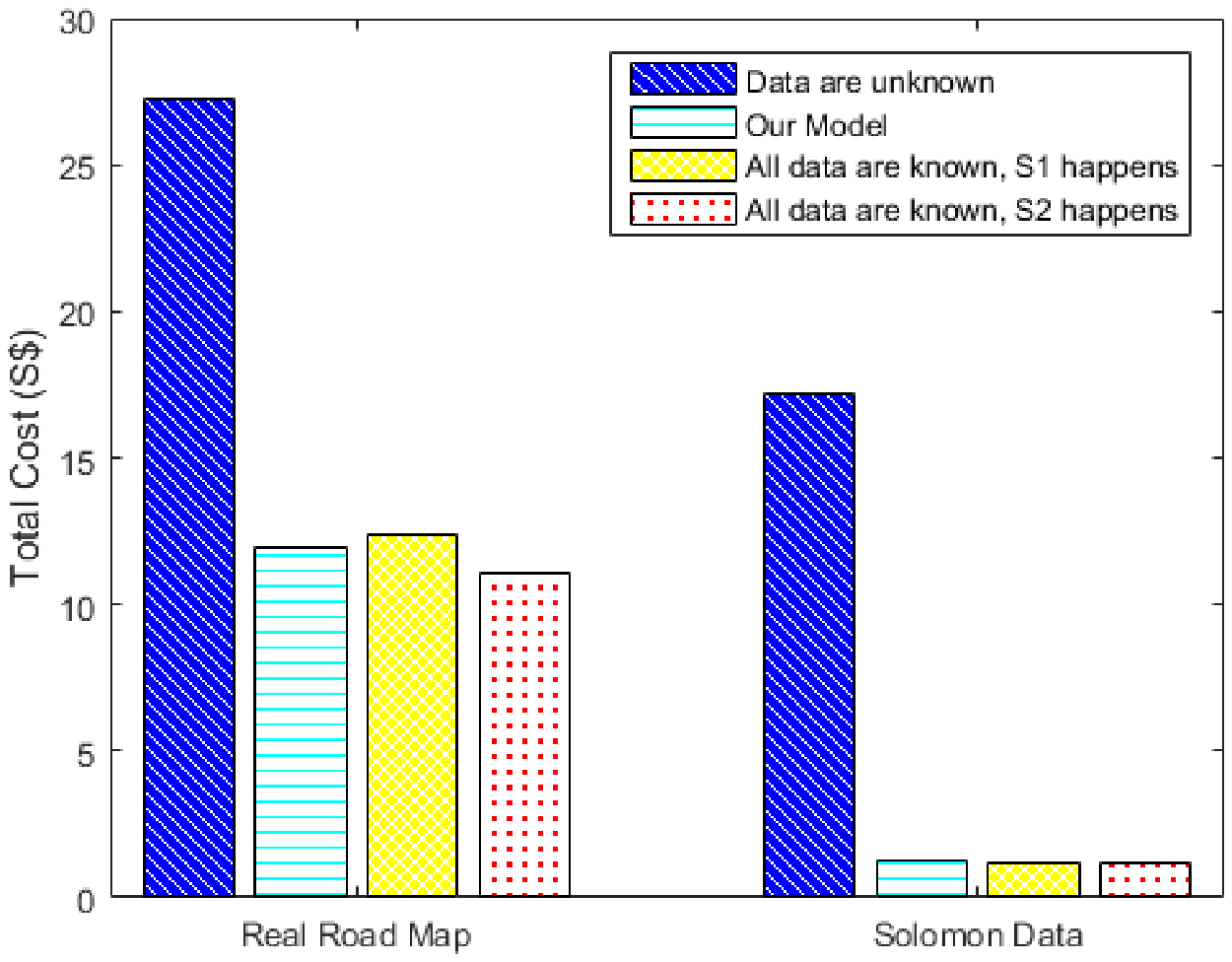}
\put(75,58) {\begin{minipage}{3em}
\fcolorbox{white}{white} {}
\end{minipage} }
\put(75,56) {\begin{minipage}{3em}
\fcolorbox{white}{white} {}
\end{minipage} }
\put(75,59) {{\tiny{$\omega_1$}}}
\put(75,56) {{\tiny{$\omega_2$}}}
\end{overpic}
\hspace{0em}
\caption{Simulation}
\label{fig_sim}
\end{minipage}
\begin{minipage}{0.66\textwidth}
$\begin{array}{cc}
\hspace{-1.5em}
\begin{overpic}[width=0.55\textwidth]{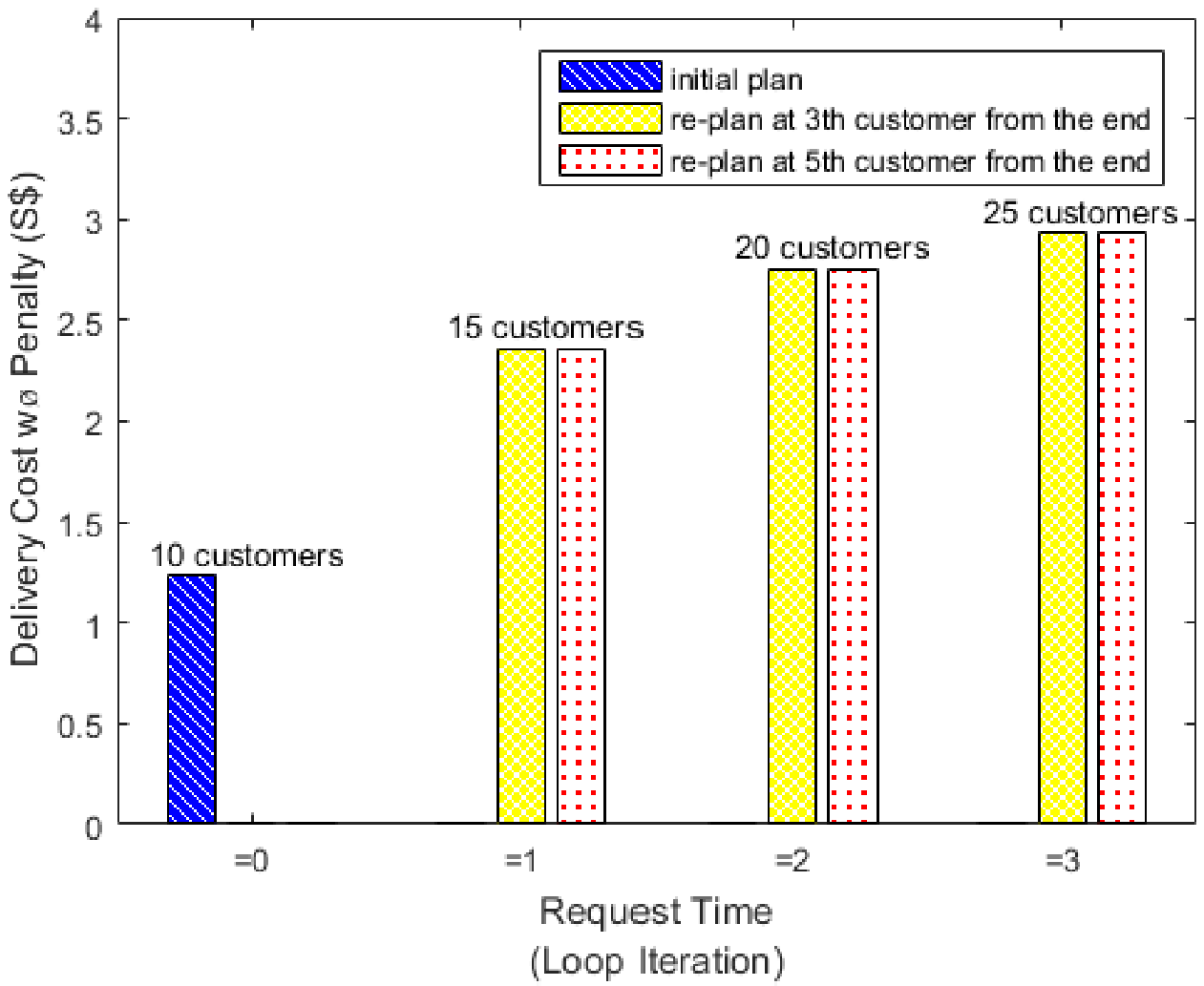}
\put(19.5,8) {\tiny$\mathbb{T}$}
\put(39,8) {\tiny$\mathbb{T}$}
\put(58.5,8) {\tiny$\mathbb{T}$}
\put(78,8) {\tiny$\mathbb{T}$}
\end{overpic}
& \hspace{-2.5em}\begin{overpic}[width=0.55\textwidth]{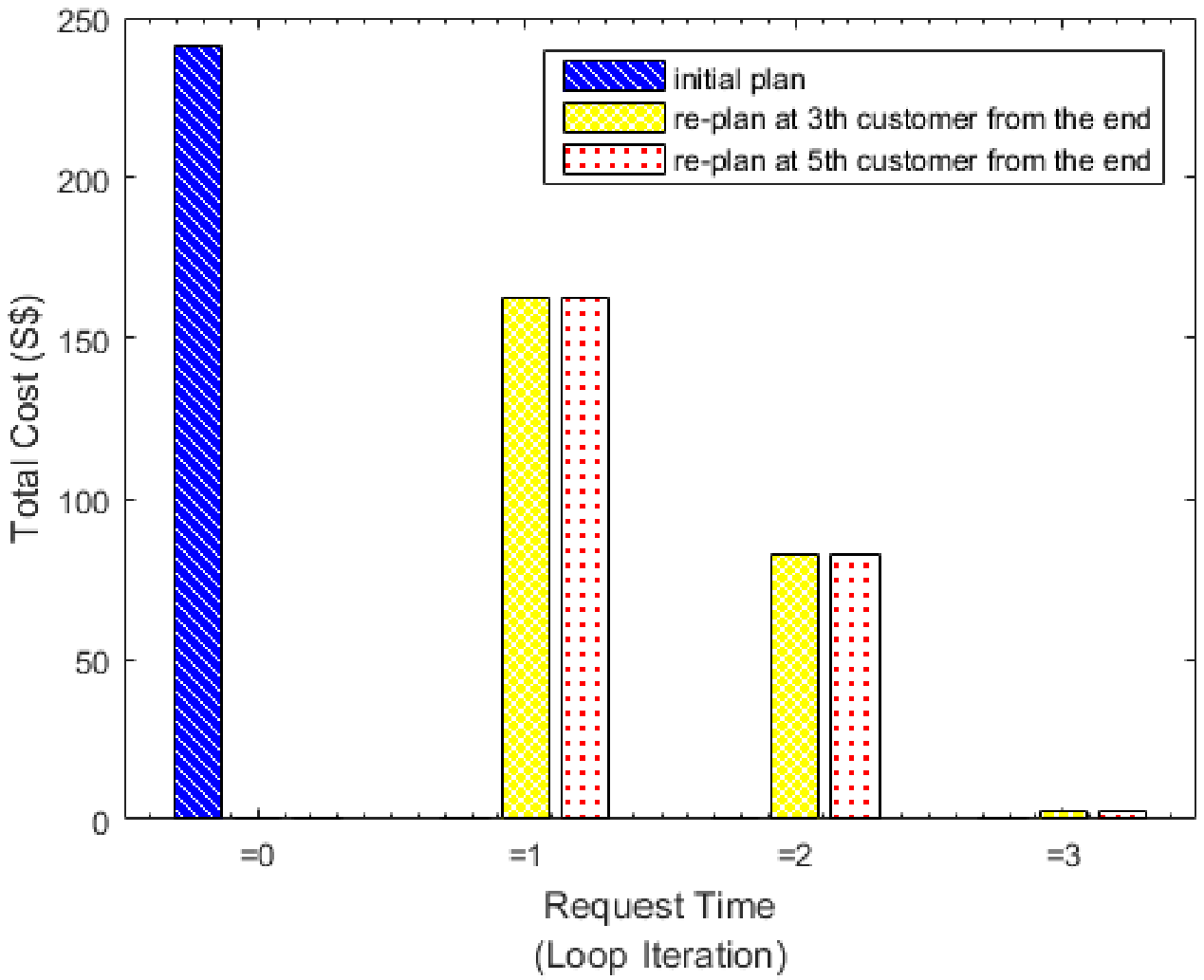}
\put(19.5,8) {\tiny$\mathbb{T}$}
\put(39,8) {\tiny$\mathbb{T}$}
\put(58.5,8) {\tiny$\mathbb{T}$}
\put(78,8) {\tiny$\mathbb{T}$}
\end{overpic}
\\
\hspace{-1.5em}(a) & \hspace{-2.5em}(b)
\end{array}$
\caption{Effectiveness of the Re-route PDPDS system, (a)  delivery cost and (b)  total delivery cost including the penalty, and thus every customers are served in (b)}
\label{fig_online}
\end{minipage}
\end{figure*}

\subsection{Impact of Stochastic model: Simulation Results}
We test the PDPSD system with ten customers from the Solomon benchmark and the real Singapore dataset. We assume that two scenarios can happen, i.e., $\Omega = \{\omega_1,\omega_2\}$, where $a_i(\omega_1) = 15$ kilograms and $a_i(\omega_2)=10$ kilograms, $\forall i \in \mathcal{C}$. All the customers request for the pickup service. However, customers $c_2$, $c_6$, and $c_{10}$ request for the delivery service, and the dependencies are set as $D_{1,2}=1$, $D_{5,6}=1$, and $D_{8,10}=1$). The simulation program is implemented in Matlab~\cite{matlab}. We compare the proposed PDPSD system, i.e., the solution of~(\ref{eq_offline_deter_obj}), with three different input settings of the deterministic system. The deterministic system is referred to as the baseline models, i.e., (i) all data are unknown, and thus the largest package weight (15 kilograms) is considered for all customers (ii) all data are known, which only scenario $\omega_1$ can happen, and (iii) all data are known, which only scenario $\omega_2$ can happen. For the PDPSD system, the probabilities are set as $P(\omega_1) = P(\omega_2) = 0.5$. 

In reality, the package sizes may not be known when the shipper plans the trips for its trucks. Figure~\ref{fig_sim} indicates that when the data are unknown, the PDPSD system achieves a much lower cost than that of the deterministic system in both datasets. Furthermore, the PDPSD system can yield marginally different total costs compared with the case that all data are known. Note that the PDPSD system achieves the lower total cost than that of the deterministic system with all data are known ($\omega_1$ happens) in the real Singapore dataset. The reason is that we simulate both $\omega_1$ and $\omega_2$ with $P(\omega_1)=P(\omega_2)=0.5$ for the PDPSD system, while only $\omega_1$ with $P(\omega_1)=1$ is used for the latter system as we assume that all data are known. Note that the total cost is higher when scenario $\omega_1$ happens than when scenario $\omega_2$ occurs. 

\subsection{Effectiveness of The Re-route PDPSD system}

We next present an impact of the re-route planning, which is presented in Section~\ref{sec_online_sto}. In this experiment, only the Solomon data is used, and only one scenario is considered. All the customers request for the pickup service with a $5$ kilograms package. We assume that the shipper owns a large truck, which can carry all customers' packages. We consider four plan and re-plan times per day ($\mathbb{T}=0,1,2,3$), and each $\mathbb{T}$ has five customers request for the service except the time $\mathbb{T}=0$ has ten customers. At time $\mathbb{T}=0$, we set $\mathcal{C} = \{c_1,\dots,c_{10}\}$. At time $\mathbb{T}=1$, we set $\mathcal{C}\cup\mathcal{F}\setminus \{\mathtt{D}\} =\{c_{1},\dots,c_{15}\}$. At time $\mathbb{T}=2$, we set $\mathcal{C}\cup\mathcal{F}\setminus \{\mathtt{D}\} =\{c_{1},\dots,c_{20}\}$. Finally, at time $\mathbb{T}=3$, we set $\mathcal{C}\cup\mathcal{F}\setminus \{\mathtt{D}\} =\{c_{1},\dots,c_{25}\}$. 

In reality, we have no information about the future requests from the customers. Therefore, the initial plan can serve only the customers at time $\mathbb{T}=0$, which are ten customers as indicated in Figure~\ref{fig_online}~(a), and thus the shipper needs to pay the outsourcing cost or penalty to meet the demand of the  customers that cannot be served by the truck. The total delivery cost including the penalty is presented in Figure~\ref{fig_online}~(b).

After the first plan, i.e., at time $\mathbb{T}=0$, the shipper can choose when to re-plan the pickup and delivery for the new customers. In this experiment, we consider the cases that the shipper re-plans its pickup and delivery when the truck arrives at the third and the fifth customers from the end of the plan. For example, the routing of the current plan is $\mathtt{D} - c_{10} - c_8 - c_9-c_6-c_2-c_4-c_1-c_3-c_5-c_7-\mathtt{D}$, and then the third and the fifth customers from the end are $c_3$ and $c_4$, respectively. Note that when re-planning at the third customer from the end, customers $c_3$, $c_5$, and $c_7$ will be considered in the re-optimization with the new customers. As shown in Figure~\ref{fig_online}, we observe that the total delivery costs are not different when the shipper re-plans the trip at the third and the fifth customers from the end because the number of different customers is too low, i.e., 2 customers.

From Figure~~\ref{fig_online}, when the shipper receives more requests, the number of serving customers increases accordingly, as well as the delivery cost without penalty because the routing cost is higher when the shipper serves more customers. In contract, to satisfy all customers, the total delivery cost decreases when the shipper accepts more requests in the Re-route PDPSD system.

\section{Conclusion}

In this paper, we have proposed the offline package pickup and delivery planning with stochastic demand (PDPSD) system and the re-route package pickup and delivery planning with stochastic demand (Re-route PDPSD) system to help a shipper effectively plans its delivery while the sizes of customer packages are random.


 The experiments have shown that when the customer package sizes are unknown, the PDPSD system can achieve a much cheaper total delivery cost than that of the deterministic system, i.e., baseline model. Moreover, the Re-route PDPSD system can help the shipper plan to serve all customers with a lower cost than that of the offline PDPSD system. For the future work, we will consider more advance formulations and techniques to covert the Re-route PDPSD system to be a fully online system. 

\section{Acknowledgment}
 {\small
This work is partially supported by Singapore Institute of Manufacturing Technology-Nanyang Technological University (SIMTech-NTU) Joint Laboratory and Collaborative research Programme on Complex Systems.}

\end{document}